\documentclass{article} %
\usepackage[table]{xcolor}    %

\usepackage{iclr2024_conference,times}

\usepackage{amsmath,amsfonts,bm}

\newcommand{\CrossQ}{Cross$Q$}

\def\eqref#1{equation~\ref{#1}}

\def\1{\bm{1}}

\def\vtheta{{\bm{\theta}}}
\def\va{{\bm{a}}}

\def\vq{{\bm{q}}}
\def\vr{{\bm{r}}}
\def\vs{{\bm{s}}}

\def\vx{{\bm{x}}}

\def\mA{{\bm{A}}}

\def\mS{{\bm{S}}}

\DeclareMathAlphabet{\mathsfit}{\encodingdefault}{\sfdefault}{m}{sl}
\SetMathAlphabet{\mathsfit}{bold}{\encodingdefault}{\sfdefault}{bx}{n}

\def\gB{{\mathcal{B}}}

\def\gL{{\mathcal{L}}}

\newcommand{\E}{\mathbb{E}}

\DeclareMathOperator*{\argmax}{arg\,max}

\usepackage[colorlinks=true,linkcolor=blue,citecolor=blue]{hyperref}
\usepackage{graphicx} 
\usepackage{amssymb}
\usepackage{url}
\usepackage{algorithm}
\usepackage{algpseudocode}
\usepackage{cancel}
\usepackage{xcolor}
\usepackage{soul} %
\usepackage{listings}
\usepackage{xcolor}
\usepackage{wrapfig}
\usepackage{cancel}
\usepackage{enumitem}
\usepackage{mathtools}
\usepackage{booktabs}

\definecolor{codegreen}{rgb}{0,0.6,0}
\definecolor{codegray}{rgb}{0.5,0.5,0.5}
\definecolor{codepurple}{rgb}{0.58,0,0.82}
\definecolor{backcolour}{rgb}{0.95,0.95,0.92}

\definecolor{tab_blue}{HTML}{1f77b4}
\definecolor{tab_orange}{HTML}{ff7f0e}
\definecolor{tab_green}{HTML}{2ca02c}
\definecolor{tab_red}{HTML}{d62728}
\definecolor{tab_purple}{HTML}{9467bd}
\definecolor{tab_brown}{HTML}{8c564b}
\definecolor{tab_pink}{HTML}{e377c2}
\definecolor{tab_gray}{HTML}{7f7f7f}
\definecolor{tab_olive}{HTML}{bcbd22}
\definecolor{tab_cyan}{HTML}{17becf}

\definecolor{fblue}{HTML}{0173b2}
\definecolor{fdarkblue}{HTML}{185a88}
\definecolor{fred}{HTML}{de8f05}
\definecolor{fgreen}{HTML}{029e73}
\definecolor{foccer}{HTML}{d55e00}
\definecolor{fmagenta}{HTML}{cc78bc}
\definecolor{fbrown}{HTML}{ca9161}
\definecolor{fpink}{HTML}{fbafe4}
\definecolor{fgray}{HTML}{949494}
\definecolor{fyellow}{HTML}{d4ca2d}
\definecolor{fcyan}{HTML}{56b4e9}

\lstdefinestyle{mystyle}{
    backgroundcolor=\color{backcolour},   
    commentstyle=\color{codegreen},
    keywordstyle=\color{magenta},
    numberstyle=\tiny\color{codegray},
    stringstyle=\color{codepurple},
    basicstyle=\ttfamily\footnotesize,
    breakatwhitespace=false,         
    breaklines=true,                 
    captionpos=b,                    
    keepspaces=true,                 
    numbers=left,                    
    numbersep=5pt,                  
    showspaces=false,                
    showstringspaces=false,
    showtabs=false,                  
    tabsize=2
}
\usepackage{array}
\newcolumntype{H}{>{\setbox0=\hbox\bgroup}c<{\egroup}@{}}

\setstcolor{red}

\algblockdefx{MRepeat}{EndRepeat}{\textbf{repeat}}{}
\algnotext{EndRepeat}

\let\oldtextbf=\textbf
\renewcommand\textbf[1]{{\boldmath\oldtextbf{#1}}}

\title{Cross$Q$: Batch Normalization \\ in Deep Reinforcement Learning \\ for Greater Sample Efficiency and Simplicity}

\makeatletter
\renewcommand*{\@fnsymbol}[1]{\ifcase#1\or*\else\@arabic{\numexpr#1-1\relax}\fi}
\makeatother

\newcommand*{\affaddr}[1]{#1} %
\newcommand*{\affmark}[1][*]{\textsuperscript{\normalfont{#1}}}

\author{%
Aditya Bhatt\affmark[~\textbf{*}~1,4]~~~~~~~~~~~~~
Daniel Palenicek\affmark[~\textbf{*}~1,2]~~~~~~~~~~~~~
Boris Belousov\affmark[~1,4]~~~~~~~~~~~~~
Max Argus\affmark[~3]\\
~~~~~~~~~~~~~~\textbf{Artemij Amiranashvili}\affmark[~3]~~~~~~~~~~~~~~
\textbf{Thomas Brox}\affmark[~3]~~~~~~~~~~~~~~
\textbf{Jan Peters}\affmark[~1,2,4,5]\\
\footnotesize
\affaddr{\affmark[\textbf{*}]Equal contribution}~
\affaddr{\affmark[1]Intelligent Autonomous Systems, TU Darmstadt}~
\affaddr{\affmark[2]Hessian.AI}~
\affaddr{\affmark[3]University of Freiburg}\\ \footnotesize
~~~~~~~~~~~~~~\affaddr{\affmark[4]German Research Center for AI (DFKI)}~~
\affaddr{\affmark[5]Centre for Cognitive Science, TU Darmstadt}\\ \footnotesize
~~~~~~~~~~~~~~~~~~~~\texttt{aditya.bhatt@dfki.de, daniel.palenicek@tu-darmstadt.de}
}

\iclrfinalcopy
\begin{document}

\maketitle
\vspace{-2em}
\begin{abstract}
Sample efficiency is a crucial problem in deep reinforcement learning.
Recent algorithms, such as REDQ and DroQ, found a way to improve the sample efficiency by increasing the update-to-data~(UTD) ratio to 20 gradient update steps on the critic per environment sample.
However, this comes at the expense of a greatly increased computational cost.
To reduce this computational burden, we introduce \CrossQ{}:
A lightweight algorithm for continuous control tasks that makes careful use of Batch Normalization and removes target networks to surpass the current state-of-the-art in sample efficiency while maintaining a low UTD ratio of $1$.
Notably, \CrossQ{} does not rely on advanced bias-reduction schemes used in current methods.
\CrossQ{}'s contributions are threefold: (1)~it matches or surpasses current state-of-the-art methods in terms of sample efficiency, (2)~it substantially reduces the computational cost compared to REDQ and DroQ, (3)~it is easy to implement, requiring just a few lines of code on top of SAC.
\end{abstract}

\section{Introduction}
\begin{wrapfigure}[27]{r}{0.42\textwidth}
    \vspace{-4.2em}
    \centering
    \includegraphics[width=0.4\textwidth]{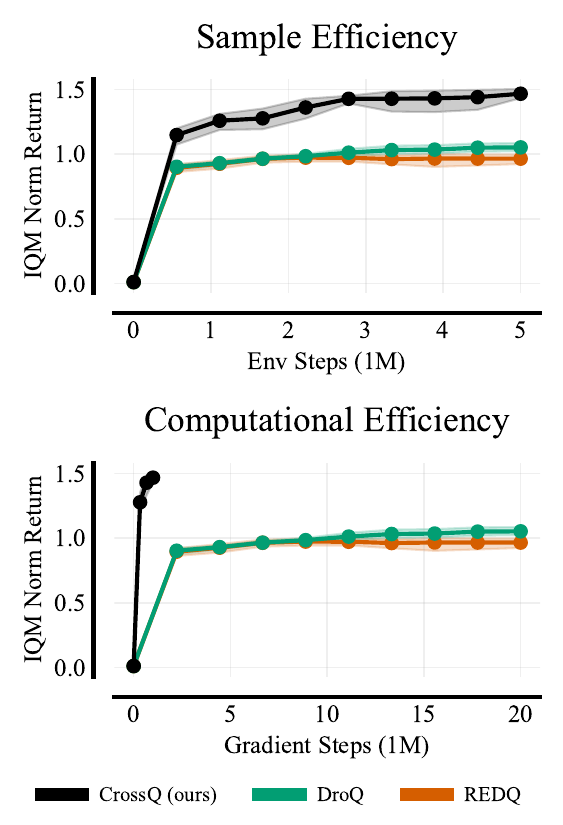}
    \vspace{-1.4em}
    \caption{\textbf{\CrossQ{} training performance aggregated over environments.} 
    \CrossQ{} is more sample efficient (top) while being significantly more computationally efficient (bottom) in terms of the gradient steps, thanks to a low UTD~$=1$.
    Following~\citet{agarwal2021rliable}, we normalize performance by the maximum of REDQ in each environment.}
    \label{fig:aggragated_efficiencies}
\end{wrapfigure}
Sample efficiency is a crucial concern when applying Deep Reinforcement Learning (Deep RL) methods on real physical systems.
One of the first successful applications of Deep RL to a challenging problem of quadruped locomotion was achieved using Soft Actor-Critic (SAC,~\citet{haarnoja2018sac}), allowing a robot dog to learn to walk within $2$h of experience~\citep{haarnoja2018soft}.
Subsequently, it was noted that the critic in SAC may be underfitted, as only a single gradient update step on the network parameters is performed for each environment step. %
Therefore, Randomized Ensembled Double Q-Learning (REDQ,~\citet{chen2021redq}) was proposed, which increased this number of gradient steps, termed update-to-data (UTD) ratio.
In addition, Dropout Q functions (DroQ,~\citet{hiraoka2021droq}) improved the computational efficiency of REDQ while maintaining the same sample efficiency by replacing its ensemble of critics with dropout. This enabled learning quadruped locomotion in a mere $20$min~\citep{smith2022walk}.
Thus, REDQ and DroQ represent the state-of-the-art in terms of sample efficiency in Deep RL for continuous control.

Importantly, both REDQ and DroQ showed that naively increasing the UTD ratio of SAC does not perform well due to the critic networks' Q value estimation bias.
Therefore, ensembling techniques were introduced for bias reduction (explicit ensemble in REDQ and implicit ensemble via dropout in DroQ), which allowed increasing the UTD to $20$ critic updates per environment step. 
Higher UTD ratios improve sample efficiency by paying the price of increased computational cost, which manifests in higher wallclock time and energy consumption.
It is, therefore, desirable to seek alternative methods that achieve the same or better sample efficiency at a 
lower computational cost, e.g., by using lower UTDs.

It turns out that even UTD~$=1$ can perform surprisingly well if other algorithmic components are adjusted appropriately.
In this paper, we introduce \textbf{\CrossQ{}}, a lightweight algorithm that achieves superior performance by removing much of the algorithmic design complexity that was added over the years, culminating in the current state-of-the-art methods.
First, it \emph{removes target networks}, an ingredient widely believed to slow down training in exchange for stability~\citep{mnih2015dqn,lillicrap2016ddpg,Kim2019DeepMellowRT,fan2020theoretical}.
Second, we find that \textit{Batch Normalization} variants (\citet{ioffe2015batchnorm};~\citet{ioffe2017batchRenorm}), when applied in a particular manner, effectively stabilize training and significantly improve sample efficiency. This contradicts others' observations that it hurts the learning performance in Deep RL, e.g.~\citet{hiraoka2021droq}.
Third, \CrossQ{} uses \textit{wider critic layers}, motivated by prior research on the ease of optimization of wider networks~\citep{ota2021widenets}.
In addition to the first two improvements, wider networks enable even higher returns.

\paragraph{Contributions.}
(1) We present the \CrossQ{} algorithm, which matches or surpasses the current state-of-the-art for model-free off-policy RL for continuous control environments with state observations in sample efficiency while being multiple times more computationally efficient;
(2)~By removing target networks, we are able to successfully accelerate off-policy Deep RL with BatchNorm;
(3) We provide empirical investigations and hypotheses for \CrossQ{}'s success.
\CrossQ{}'s changes mainly pertain to the deep network architecture of SAC; therefore, our study is chiefly empirical: through a series of ablations, we isolate and study the contributions of each part.
We find that \CrossQ{} matches or surpasses the state-of-the-art algorithms in sample efficiency while being up to $4\times$ faster in terms of wallclock time without requiring critic ensembles, target networks, or high UTD ratios.
We provide the \CrossQ{} source code at~\href{https://github.com/adityab/CrossQ}{github.com/adityab/CrossQ}.

\section{Background}

\subsection{Off-policy Reinforcement Learning and Soft Actor-Critic}
We consider a discrete-time Markov Decision Process~(MDP,~\citet{puterman2014mdp}), defined by the tuple $\langle\mathcal{S}, \mathcal{A}, \mathcal{P}, \mathcal{R}, \rho, \gamma\rangle$
with state space $\mathcal{S}$, action space $\mathcal{A}$, 
transition probability $\vs_{t+1}\sim\mathcal{P}(\cdot|\vs_t,\va_t)$, 
reward function $r_t = \mathcal{R}(\vs_t,\va_t)$,
initial state distribution $\vs_0\sim \rho$
and discount factor $\gamma \in [0, 1)$.
RL describes the problem of an agent learning an optimal policy $\pi$ for a given MDP.
At each time step $t$, the agent receives a state $\vs_t$ and interacts with the environment according to its policy $\pi$.
We focus on the Maximum Entropy RL setting~\citep{ziebart2008maxent}, where the agent's objective is to find the optimal policy $\pi^*$, which maximizes the expected cumulative reward while keeping the entropy~$\mathcal{H}$ high;
$
    \textstyle \argmax_{\pi^*} \E_{\vs_0 \sim \rho} 
    \left[ \sum_{t=0}^{\infty} \gamma^t (r_t - \alpha \mathcal{H}(\pi(\:\cdot\:|\vs_t))) \right].
$
The action-value function is defined by
$
    Q(\vs,\va)=\mathbb{E}_{\pi,\mathcal{P}}\left[\sum_{t=0}^\infty \gamma^t ( r_t - \alpha \log \pi(\va_t | \vs_{t}) )|\vs_0=\vs,\va_0=\va\right]
$
and describes the expected reward when taking action $\va$ in state $\vs$.
Soft Actor-Critic~(SAC,~\citep{haarnoja2018sac}) is a popular algorithm that solves the MaxEnt RL problem.
SAC parametrizes the Q function and policy as neural networks and trains two independent versions of the Q function, using the minimum of their estimates to compute the regression targets for Temporal Difference (TD) learning.
This \textit{clipped double-Q} trick, originally proposed by~\citet{fujimoto2018td3} in TD3, helps in reducing the potentially destabilizing overestimation bias inherent in approximate Q-learning~\citep{hasselt2010double}.

\subsection{High update-to-data Ratios, REDQ, and DroQ}
Despite its popularity among practitioners and as a foundation for other more complex algorithms, SAC leaves much room for improvement in terms of sample efficiency.
Notably, SAC performs exactly one gradient-based optimization step per environment interaction.
SAC's UTD~$=1$ setting is analogous to simply training for fewer epochs in supervised learning.
Therefore, in recent years, gains in sample efficiency within RL have been achieved through increasing the UTD ratio~\citep{janner2019mbpo,chen2021redq,hiraoka2021droq,nikishin2022primacy}.
Different algorithms, however, substantially vary in their approaches to achieving high UTD ratios.
\citet{janner2019mbpo} uses a model to generate synthetic data, which allows for more overall gradient steps. 
\citet{nikishin2022primacy} adopt a simpler approach: they increase the number of gradient steps while periodically resetting the policy and critic networks to fight premature convergence to local minima. We now briefly outline the two high-UTD methods to which we compare \CrossQ{}.

\vspace{-0.5em}
\paragraph{REDQ.} \citet{chen2021redq} find that merely raising SAC's UTD ratio hurts performance. They attribute this to the accumulation of the learned Q functions' estimation bias over multiple update steps---despite the clipped double-Q trick---which destabilizes learning. To remedy this bias more strongly, they increase the number of Q networks from two to an ensemble of 10. Their method, called REDQ, permits stable training at high UTD ratios up to 20.

\vspace{-0.5em}
\paragraph{DroQ.} \citet{hiraoka2021droq} note that REDQ's ensemble size, along with its high UTD ratio, makes training computationally expensive. They instead propose using a smaller ensemble of Q functions equipped with Dropout~\citep{srivastava2014dropout}, along with Layer Normalization~\citep{ba2016layernorm} to stabilize training in response to the noise introduced by Dropout. Called DroQ, their method is computationally cheaper than REDQ, yet still expensive due to its UTD ratio of 20.

\section{The Cross$Q$ Algorithm}
\begin{figure}[t]
    \begin{lstlisting}[language=Python]
def critic_loss(Q_params, policy_params, obs, acts, rews, next_obs):
    next_acts, next_logpi = policy.apply(policy_params, obs)

    # Concatenated forward pass
    all_q, new_Q_params = Q.apply(Q_params,
        jnp.concatenate([obs, next_obs]), 
        jnp.concatenate([acts, next_acts])
    )
    # Split all_q predictions and stop gradient on next_q
    q, next_q = jnp.split(all_q, 2)
    next_q = jnp.min(next_q, axis=0)   # min over double Q function
    next_q = jax.lax.stop_gradient(next_q - alpha * next_logpi)
    return jnp.mean((q - (rews + gamma * next_q))**2), new_Q_params\end{lstlisting}
    \vspace{-1.1em}
    \caption{\textbf{\CrossQ{} critic loss in JAX.} The \CrossQ{} critic loss is easy to implement on top of an existing SAC implementation. One just adds the batch normalization layers into the critic network and removes the target network. As we are now left with only the critic network, one can simply concatenate observations and next observations, as well as actions and next actions along the batch dimension, perform a joint forward pass, and split up the batches afterward.
    Combining two forward passes into one grants a small speed-up thanks to requiring only one CUDA call instead of two.
    }
    \label{fig:crossq_python_code}
\end{figure}
In this paper, we challenge this current trend of high UTD ratios and demonstrate that we can achieve competitive sample efficiency at a much lower computational cost with a UTD~$=1$ method.
\CrossQ{} is our new state-of-the-art off-policy actor-critic algorithm.
Based on SAC, it uses purely network-architectural
engineering insights from deep learning to accelerate training. 
As a result, it \xcancel{crosses out} much of the algorithmic design complexity that was added over the years and which led to the current state-of-the-art methods. In doing so, we present a much simpler yet more efficient algorithm. 
In the following paragraphs, we introduce the three design choices that constitute \CrossQ{}.

\subsection{Design Choice 1: Removing Target Networks}
\citet{mnih2015dqn} originally introduced target networks to stabilize the training of value-based off-policy RL methods, and today, most algorithms require them~\citep{lillicrap2016ddpg,fujimoto2018td3,haarnoja2018sac}. SAC updates the critics' target networks with Polyak Averaging
\begin{align}
\label{eq:taget_polyak}
    \textstyle \vtheta^\circ \leftarrow (1-\tau) \vtheta^\circ + \tau \vtheta,
\end{align}
where $\vtheta^\circ$ are the target network parameters, and $\vtheta$ are those of the trained critic. Here $\tau$ is the \textit{target network smoothing coefficient}; with a high $\tau=1$ (equivalent to cutting out the target network), SAC training can diverge, leading to explosive growth in $\vtheta$ and the $Q$ predictions.
Target networks stabilize training by explicitly delaying value function updates, arguably slowing down online learning~\citep{plappert2018multi, Kim2019DeepMellowRT,morales2020grokking}.

Recently,~\citet{yang2021overcoming} found that critics with Random Fourier Features can be trained without target networks, suggesting that the choice of layer activations affects the stability of training. Our experiments in Section~\ref{sec:ablation} uncover an even simpler possibility: using bounded activation functions or feature normalizers is sufficient to prevent critic divergence in the absence of target networks, whereas the common choice of \texttt{relu} without normalization diverges. While others have used normalizers in Deep RL before, we are the first to identify that they make target networks redundant. Our next design choice exploits this insight to obtain an even greater boost.

\subsection{Design Choice 2: Using Batch Normalization}
\label{sec:design_choice_BN}
\begin{figure}[t]
\begin{tabular}{@{\hspace{2cm}}l@{\hspace{1cm}}l}
    \multicolumn{1}{c}{SAC:} & \multicolumn{1}{c}{\CrossQ{} (Ours):} \\[8pt]
    
    {$\!\begin{aligned}
        {\color{magenta}Q_\vtheta}(\mS_t,\mA_t) &= \vq_t \\
        {\color{magenta}Q_{\vtheta^\circ}}(\mS_{t+1},\mA_{t+1}) &= {\color{purple}\vq_{t+1}^\circ}
    \end{aligned}$} & 
    {$\!\begin{aligned}  
        {\color{cyan}Q_\vtheta}\left(
        \begin{bmatrix}
            \begin{aligned}  
                &\mS_t \\
                &\mS_{t+1}
            \end{aligned}
        \end{bmatrix},
        \begin{bmatrix}
            \begin{aligned} 
                &\mA_t \\
                &\mA_{t+1}
            \end{aligned}
        \end{bmatrix}
        \right) = 
        \begin{bmatrix}
            \begin{aligned} 
                &\vq_t \\ 
                &\vq_{t+1} 
            \end{aligned}
        \end{bmatrix}
    \end{aligned}$} \\[15pt]
    {$\!\begin{aligned}
        \gL_{\color{magenta}\vtheta} &= (\vq_t - \vr_t - \gamma\, {\color{purple}\vq^\circ_{t+1}})^2
    \end{aligned}$} &
    {$\!\begin{aligned}
        \gL_{\color{cyan}\vtheta} &= (\vq_t - \vr_t - \gamma\,|\vq_{t+1}|_{\mathtt{sg}})^2
    \end{aligned}$}
\end{tabular}
\caption{SAC \textcolor{magenta}{without BatchNorm in the critic} ${\color{magenta}Q_\vtheta}$ (left) requires \textcolor{purple}{target $Q$ values $\vq_{t+1}^\circ$} to stabilize learning. 
\CrossQ{} \textcolor{cyan}{with BatchNorm in the critic} ${\color{cyan}Q_\vtheta}$ (right) removes the need for target networks and allows for a joint forward pass of both current and future values.
Batches are sampled from the replay buffer $\gB$: $\mS_t, \mA_t, \vr_t, \mS_{t+1}\sim \gB$ and $\mA_{t+1}\sim \pi_\phi(\mS_{t+1})$ from the current policy.
$|\cdot|_{\mathtt{sg}}$ denotes the \texttt{stop-gradient} operation.
}
\label{fig:jointForwardPass}
\end{figure}

\begin{wrapfigure}[18]{r}{0.3\textwidth}
    \centering
    \vspace{-5.3em}
    \includegraphics[width=0.27\textwidth]{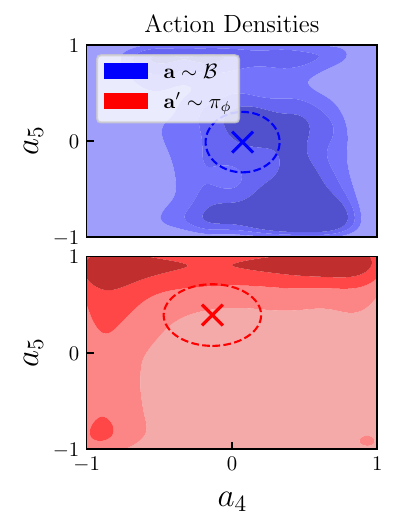}
    \vspace{-1.4em}
    \caption{\textbf{Replay buffer and current policy actions are distributed differently.} Darker colors denote higher density. Estimated from a batch of $10^4$ transitions $(\va,\vs')\sim\mathcal{B};$ $\va'\sim\pi_\phi(\vs')$, after $3\times10^5$ training steps on \texttt{Walker2d}; $a_4$ and $a_5$ are random action dimensions.}
    \label{ref:action_densities}
\end{wrapfigure}

BatchNorm has not yet seen wide adoption in value-based off-policy RL methods, despite its success and widespread use in supervised learning~\citep{he2016resnet, santurkar2018howbatchnorm}, attempts at doing so have fared poorly. \citet{lillicrap2016ddpg} use BatchNorm layers on the state-only representation layers in the DDPG critic but find that it does not help significantly. Others use BatchNorm in decoupled feature extractors for Deep RL networks~\citep{ota2020can, ota2021widenets}, but not in critic networks. \citet{hiraoka2021droq} report that using BatchNorm in critics causes training to fail in DroQ.

\hspace{0em}\textbf{We find using BatchNorm \emph{carefully}, when \emph{additionally} removing target networks, performs surprisingly well, trains stably, and is, in fact, algorithmically simpler than current methods.}

First, we explain why BatchNorm needs to be used \emph{carefully}.
Within the critic loss $[Q_\vtheta(\mS,\mA) - (\vr + \gamma Q_{\vtheta^\circ}(\mS',\mA'))]^2$, predictions are made for two differently distributed batches of state-action pairs; $(\mS,\mA)$ and $(\mS',\mA')$, where $\mA'\sim\pi_\phi(\mS')$ is sampled from the \textit{current policy}, while $\mA$ originates from old behavior policies.

Just like the target network, the BatchNorm parameters are updated by Polyak Averaging from the live network~(Equation~\ref{eq:taget_polyak}).
The BatchNorm running statistics of the live network, which were estimated from batches of $(\vs,\va)$ pairs, will clearly not have \textit{seen} samples $(\vs',\pi_\phi(\vs'))$ and will further not match their statistics.
In other words, the state-action inputs evaluated by the target network will be out-of-distribution, given its mismatched BatchNorm running statistics.
It is well known that the prediction quality of BatchNorm-equipped networks degrades in the face of such test-time distribution shifts~\citep{pham2022continual, lim2023ttn}.

Removing the target network provides an \textit{elegant} solution.
With the target network removed, we can concatenate both batches and feed them through the $Q$ network in a single forward pass, as illustrated in Figure~\ref{fig:jointForwardPass} and shown in code in Figure~\ref{fig:crossq_python_code}. This simple trick ensures that BatchNorm's normalization moments arise from the union of both batches, corresponding to a $50/50$ mixture of their respective distributions. Such normalization layers \textit{do not} perceive the $(\vs',\pi_\phi(\vs'))$ batch as being out-of-distribution. This small change to SAC allows the safe use of BatchNorm and greatly accelerates training. 
We are not the only ones to identify this way of using BatchNorm to tackle the distribution mismatch; other works in supervised learning, e.g., 
Test-Time Adaptation \citep{lim2023ttn}, EvalNorm \citep{singh2019evalnorm}, and \textit{Four Things Everyone Should Know to Improve Batch Normalization} \citep{Summers2020Four} also use mixed moments to bridge this gap.

In practice, \CrossQ{}'s actor and critic networks use Batch Renormalization~(\texttt{BRN},~\citet{ioffe2017batchRenorm}), an improved version of the original \texttt{BN}~\citep{ioffe2015batchnorm} that is robust to long-term training instabilities originating from minibatch noise. \texttt{BRN} performs batch normalization using the less noisy \textit{running statistics} after a warm-up period, instead of noisy minibatch estimates as in \texttt{BN}. In the rest of this paper, all discussions with ``BatchNorm'' apply equally to both versions unless explicitly disambiguated by \texttt{BN} or \texttt{BRN}.

\subsection{Design Choice 3: Wider Critic Networks}
Following \citet{ota2021widenets}, we find that wider critic network layers in \CrossQ{} lead to even faster learning.
As we show in our ablations in Section~\ref{sec:ablation}, most performance gains originate from the first two design choices; however, wider critic networks further boost the performance, helping to match or outperform REDQ and DroQ sample efficiency.

\begin{figure}[t]
    \centering
    \includegraphics[width=\textwidth]{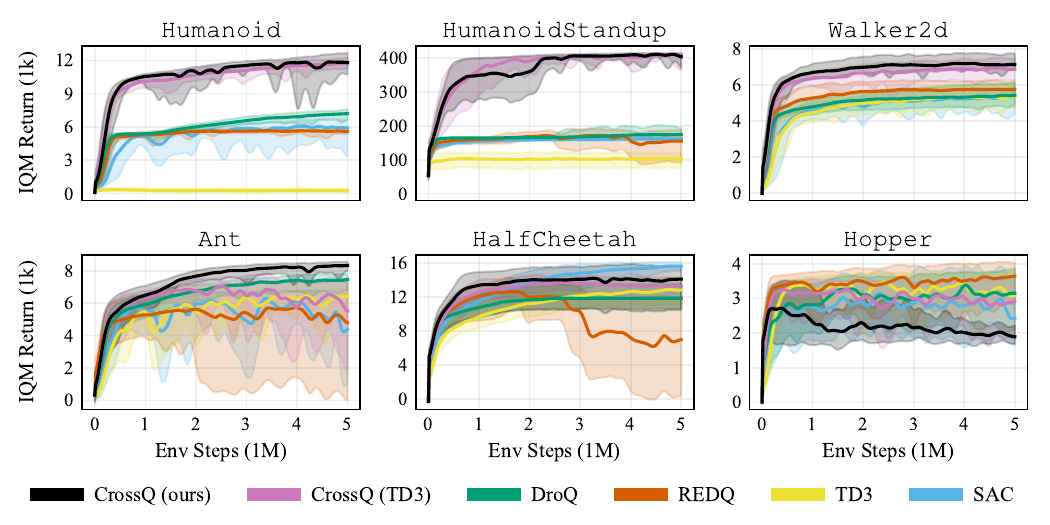}
    \vspace{-2em}
    \caption{\textbf{\CrossQ{} sample efficiency.}
    Compared to REDQ and DroQ (UTD $=20$) \CrossQ{} (UTD $=1$) performs either comparably, better, or---for the more challenging \texttt{Humanoid} tasks---substantially better. These results directly transfer to TD3 as the base algorithm in \CrossQ{} (TD3). We plot \textit{interquartile mean} (IQM) and $70\%$ quantile interval of the episodic returns over $10$ seeds. 
    }
    \vspace{-1em}
    \label{ref:sample_efficiency}
\end{figure}

We want to stress again that \textbf{\CrossQ{}}, a UTD $=1$ method, \textbf{\textit{does not use bias-reducing ensembles, high UTD ratios or target networks}}. Despite this, it achieves its competitive sample efficiency at a fraction of the compute cost of REDQ and DroQ (see Figures~\ref{ref:sample_efficiency} and~\ref{ref:compute_efficiency}). Note that our proposed changes can just as well be combined with other off-policy TD-learning methods, such as TD3, as shown in our experiments in Section~\ref{sec:sample_efficiency}.

\section{Experiments and Analysis}
We conduct experiments to provide empirical evidence for \CrossQ{}'s performance,
and investigate:
\vspace{-0.5em}
\begin{enumerate}[noitemsep]
    \item Sample efficiency of \CrossQ{} compared to REDQ and DroQ;%
    \item Computational efficiency in terms of wallclock time and performed gradient step;%
    \item Effects of the proposed design choices on the performance via Q function bias evaluations;
\end{enumerate}
\vspace{-0.5em}
And conduct further ablation studies for the above design choices. We evaluate across a wide range of continuous-control \texttt{MuJoCo}~\citep{todorov2012mujoco} environments, with $10$ random seeds each. Following~\citet{janner2019mbpo,chen2021redq} and~\citet{hiraoka2021droq}, we evaluate on the same four \texttt{Hopper}, \texttt{Walker2d}, \texttt{Ant}, and \texttt{Humanoid} tasks, as well as two additional tasks: \texttt{HalfCheetah} and the more challenging \texttt{HumanoidStandup} from Gymnasium~\citep{towers2023gymnasium}.
We adapted the JAX version of stable-baselines~\citep{stable-baselines3} for our experiments.

\subsection{Sample Efficiency of \CrossQ{}}
\label{sec:sample_efficiency}
Figure~\ref{ref:sample_efficiency} compares our proposed \CrossQ{} algorithm with REDQ, DroQ, SAC and TD3 in terms of their sample efficiency, i.e., average episode return at a given number of environment interactions. As a proof of concept, we also present \CrossQ{}~(TD3), a version of \CrossQ{} which uses TD3 instead of SAC as the base algorithm.
We perform periodic evaluations during training to obtain the episodic reward. From these, we report the mean and standard deviations over $10$ random seeds.
All subsequent experiments in this paper follow the same protocol.

This experiment shows that \CrossQ{} matches or outperforms the best baseline in all the presented environments except on \texttt{Ant}, where REDQ performs better in the early training stage, but \CrossQ{} eventually matches it.
On \texttt{Hopper}, \texttt{Walker}, and \texttt{HalfCheetah}, the learning curves of \CrossQ{} and REDQ overlap, and there is no significant difference.
On the harder \texttt{Humanoid} and \texttt{HumanoidStandup} tasks, \CrossQ{} and \CrossQ{}~(TD3) both substantially surpass all baselines.

\subsection{Computational Efficiency of \CrossQ{}}
\label{sec:computational_efficiency}
\begin{figure}[t]
    \centering
    \includegraphics[width=\textwidth]{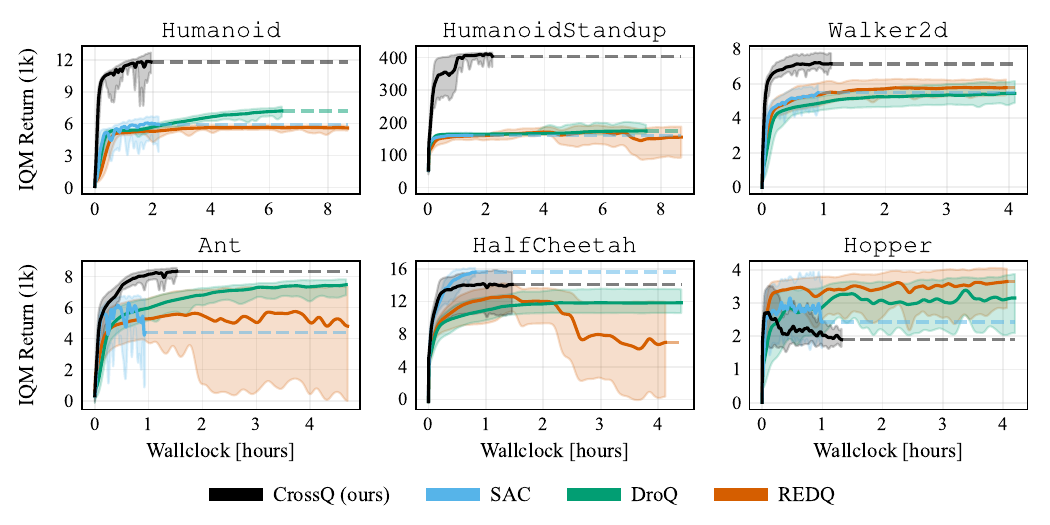}
    \vspace{-2em}
    \caption{\textbf{Computational efficiency.} \CrossQ{} trains an order of magnitude faster, taking only $5\%$ of the gradient steps, substantially saving on wallclock time.
    The dashed horizontal lines are visual aids to better compare the final performance after training for $5\times10^6$ environment steps. We plot IQM and $70\%$ quantile interval over $10$ seeds. Appendix~\ref{sec:wallclock} provides a table of wallclock times.}
    \label{ref:compute_efficiency}
    \vspace{-1em}
\end{figure}

Figure~\ref{ref:compute_efficiency} compares the computational efficiency of \CrossQ{} to the baselines.
This metric is where \CrossQ{} makes the biggest leap forward. \CrossQ{} requires $20\times$ fewer gradient steps than REDQ and DroQ, which results in roughly $4\times$ faster wallclock speeds (Table~\ref{tab:computation_times}).
Especially on the more challenging \texttt{Humanoid} and \texttt{HumanoidStandup} tasks the speedup is the most pronounced.
In our view, this is a noteworthy feature.
On the one hand, it opens the possibility of training agents in a truly online and data-efficient manner, such as in real-time robot learning.
On the other hand, with large computing budgets \CrossQ{} can allow the training of even larger models for longer than what is currently feasible, because of its computational efficiency stemming from its low UTD~$=1$.

\subsection{Evaluating $Q$ Function Estimation Bias}
\label{sec:q_bias}
\begin{figure}[t]
\label{fig:q_bias_2envs}
    \centering
    \includegraphics[width=\textwidth]{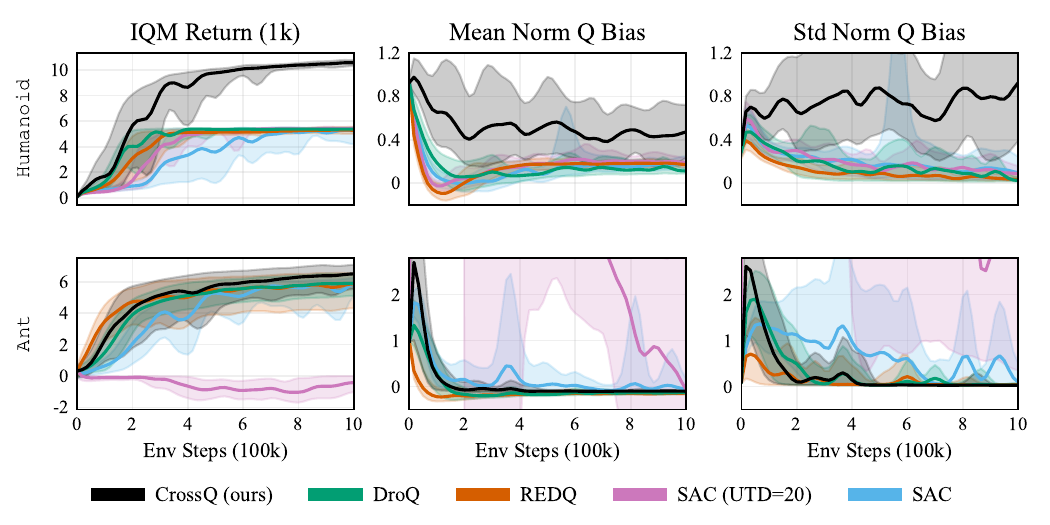}
    \vspace{-2em}
    \caption{\textbf{$Q$ estimation bias does not reliably influence learning performance}. Following the analysis of~\citet{chen2021redq}, we plot the IQM and $70\%$ quantile interval of the normalized Q function bias. REDQ generally has the least bias over $10$ seeds. \CrossQ{} matches or outperforms DroQ, REDQ and SAC while showing more Q function bias in all environments. The full set of environments is shown in Fig.~\ref{fig:q_bias} in the Appendix.}
    \label{fig:q_bias}
    \vspace{-1em}
\end{figure}

All methods we consider in this paper are based on SAC and, thus, include the clipped double-Q trick to reduce Q function overestimation bias~\citep{fujimoto2018td3}. \citet{chen2021redq} and \citet{hiraoka2021droq} stress the importance of keeping this bias even lower to achieve their high performances and intentionally design REDQ and DroQ to additionally reduce bias with explicit and implicit ensembling. In contrast, \CrossQ{} outperforms both baselines without any ensembling. Could \CrossQ{}'s high performance be attributed to implicitly reducing the bias as a side effect of our design choices? Using the same evaluation protocol as~\citet{chen2021redq}, we compare the normalized Q prediction biases in Figure~\ref{fig:q_bias_2envs}. Due to space constraints, here we show \texttt{Hopper} and \texttt{Ant} and place the rest of the environments in Figure~\ref{fig:q_bias} in the Appendix.

We find that REDQ and DroQ indeed have lower bias than SAC and significantly lower bias than SAC with UTD $=20$. The results for \CrossQ{} are mixed: while its bias trend exhibits a lower mean and variance than SAC, in some environments, its bias is higher than DroQ, and in others, it is lower or comparable. REDQ achieves comparable or worse returns than CrossQ while maintaining the least bias. As \CrossQ{} performs better \textit{despite} having---perhaps paradoxically---generally higher Q estimation bias, we conclude that the relationship between performance and estimation bias is complex, and one does not seem to have clear implications on the other.

\subsection{Ablations}
\label{sec:ablation}
We conduct ablation studies to better understand the impact of different design choices in \CrossQ{}.

\subsubsection{\small{Disentangling the Effects of Target Networks and BatchNorm}}
\label{sec:bounded_activations}
\begin{figure}[t]
    \centering
    \includegraphics[width=\textwidth]{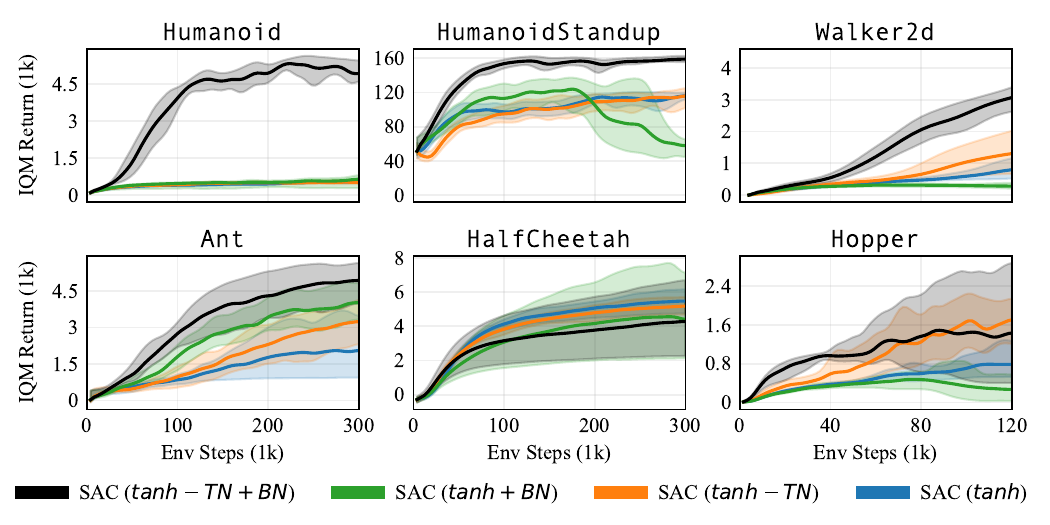}
    \vspace{-2em}
    \caption{\textbf{The effects of target networks and BatchNorm on sample efficiency.} All SAC variants in this experiment use critics with $\mathrm{tanh}$ activations, since they allow divergence-free training without target networks, enabling this comparison. This ablation uses the original BatchNorm (\texttt{BN}, \citet{ioffe2015batchnorm}). Removing target networks (\texttt{-TN}) provides only small improvements over the SAC baseline with target nets. BatchNorm with target nets (\texttt{+BN}, green) is unstable. Using BatchNorm after removing target nets (\texttt{-TN+BN})---the configuration most similar to \CrossQ{}---performs best. We plots IQM return and $70\%$ quantile intervals over $10$ seeds.}
    \vspace{-1em}
\label{fig:bounded_activations}
\end{figure}

\CrossQ{} changes SAC in three ways; of these, two explicitly aim to accelerate optimization: the removal of target networks, and the introduction of BatchNorm. Unfortunately, SAC without target networks diverges; therefore, to study the contribution of the first change, we need a way to compare SAC---divergence-free---\textit{with and without target networks}. Fortunately, we find that such a way exists: according to our supplementary experiments in Appendix~\ref{sec:diverse_activations}, simply using bounded activation functions in the critic appears to prevent divergence. This is a purely empirical observation and an in-depth study regarding the influence of activations and normalizers on the stability of Deep RL is beyond the scope of this paper. In this specific ablation, we use \texttt{tanh} activations instead of \texttt{relu}, solely as a tool to make the intended comparison possible.

Figure~\ref{fig:bounded_activations} shows the results of our experiment. The performance of SAC without target networks supports the common intuition that target networks indeed slow down learning to a small extent. We find that the combination of BatchNorm and Target Networks performs inconsistently, failing to learn anything in half of the environments. Lastly, the configuration of BatchNorm without target networks---and the closest to \CrossQ{}---achieves the best aggregate performance, with the boost being significantly bigger than that from removing target networks alone.
In summary, even though removing target networks may slightly improve performance in some environments, it is the combination of removing target networks and adding BatchNorm that accelerates learning the most.

\subsubsection{Ablating the Different Design Choices and Hyperparameters}
\begin{wrapfigure}[20]{tr}{0.45\textwidth}
    \vspace{-1.5em}
    \centering
    \includegraphics[width=0.45\textwidth]{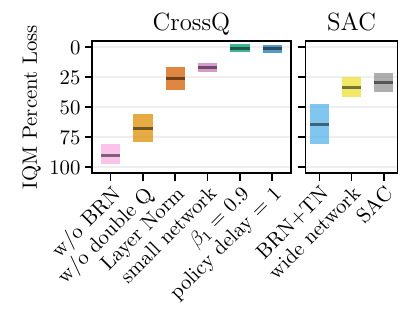}
    \vspace{-2.45em}
    \caption{\textbf{Ablations on \CrossQ{} and SAC.}
    Loss in IQM return in percent---relative to \CrossQ{}---at $1$M environment interactions. Aggregated over all environments and six seeds each, with $95\%$ bootstrapped confidence intervals~\citep{agarwal2021rliable}.
    Left shows \CrossQ{} ablations; Right shows effects of adding parts on top of SAC.
    Figure~\ref{fig:crossq_ablations} in Appendix shows individual training curves.
    }
    \vspace{-1em}
\label{fig:bar_ablations}
\end{wrapfigure}

\begin{figure}[t]
    \centering
    \includegraphics[width=\textwidth]{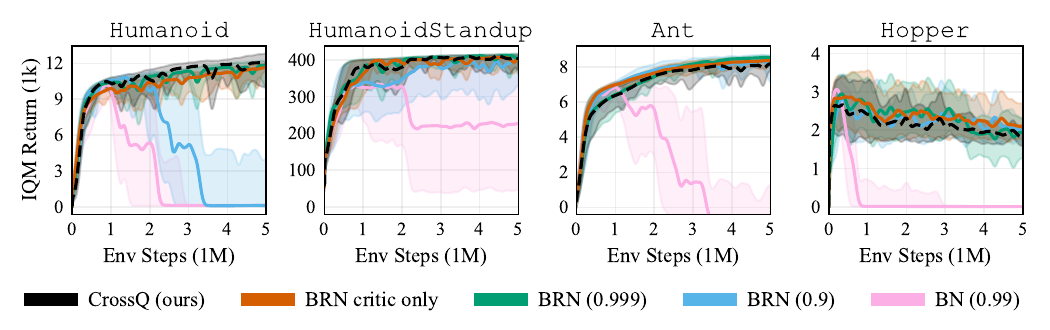}
    \vspace{-2em}
    \caption{\textbf{Comparing BatchNorm hyperparameters.} All variants have comparably strong and stable curves early in the training.
    Omitting normalization in the actor (\texttt{BRN} critic only) does not significantly affect \CrossQ{}. Using the original Batch Normalization (\texttt{BN}, with moving-average momentum $0.99$) is prone to sudden performance collapses during longer training runs. Using \texttt{BRN} permits stabler training, which improves with higher momentums; \CrossQ{}'s default $0.99$~(black) and higher show no collapses.
    We plot IQM return and $70\%$ quantile intervals over five seeds.}
    \vspace{-1em}
\label{fig:batchnorm_ablations}
\end{figure}

In this subsection, we examine the contributions of the different \CrossQ{} design choices to show their importance.
Figure~\ref{fig:bar_ablations} shows aggregated ablations of these components and various hyperparameters, while Figure~\ref{fig:batchnorm_ablations} ablates the BatchNorm layer itself.

\paragraph{Hyperparameters.}
\CrossQ{} uses the best hyperparameters obtained from a series of grid searches. Of these, only three are different from SAC's default values. 
First, we find that \textcolor{fgreen}{reducing the~$\beta_1$ momentum} for the Adam optimizer~\citep{Kingma2014AdamAM} from $0.9$ to $0.5$ as well the \textcolor{fdarkblue}{\textit{policy delay} of $3$} have the smallest impact on the performance. However, since fewer actor gradient steps reduce compute, this setting is favorable.
Second, \textcolor{fmagenta}{reducing the critic network's width to 256}---the same small size as SAC---reduces performance and yet still significantly outperforms SAC.
This suggests that practitioners may be able to make use of a larger compute budget, i.e., train efficiently across a range of different network sizes, by scaling up layer widths according to the available hardware resources.
Third, as expected, \textcolor{fpink}{removing the \texttt{BRN} layers} proves to be detrimental and results in the worst overall performance. 
A natural question that comes to mind is whether other normalization strategies in the critic, such as Layer Normalization (LayerNorm,~\citet{ba2016layernorm}), would also give the same results. However, in our ablation, we find that \textcolor{foccer}{replacing BatchNorm with LayerNorm} degrades \CrossQ{}'s performance significantly, roughly to the level of the SAC baseline.
Lastly, SAC does not benefit from simply \textcolor{fyellow}{widening critic layers to $2048$}. 
And \textcolor{fblue}{naively adding \texttt{BRN} to SAC while keeping the target networks} proves detrimental. This finding is in line with our diagnosis of mismatched statistics being detrimental to the training.

\paragraph{Batch Normalization Layers.}
 
In Figure \ref{fig:batchnorm_ablations}, we ablate
the BatchNorm versions (\texttt{BN}~\citep{ioffe2015batchnorm} and \texttt{BRN}~\citep{ioffe2017batchRenorm}) and their internal moving-average momentums. Compared to \CrossQ{}'s optimal combination---\texttt{BRN} with momentum $0.99$---all variants have similar sample efficiency in the early stages of training (1M steps). When using \texttt{BN}, we sometimes observe sudden performance collapses later in training; we attribute these to \texttt{BN}'s unique approach of using noisy \textit{minibatch estimates} of normalization moments. \texttt{BRN}'s improved approach of using the less noisy \textit{moving-averages} makes these collapses less likely; further noise-reduction via higher momentums eliminates these collapses entirely. Additionally, we find that using BatchNorm only in the critic (instead of both the actor and the critic) is sufficient to drive the strong performance of \CrossQ{}; however, including it in both networks performs slightly better.

\section{Conclusion \& Future Work}
We introduced \CrossQ{}, a new off-policy RL algorithm that matches or exceeds the performance of REDQ and DroQ---the current state-of-the-art on continuous control environments with state observations---in terms of sample efficiency while being multiple times more computationally efficient.
To the best of our knowledge, \CrossQ{} is the first method to successfully use BatchNorm to greatly accelerate off-policy actor-critic RL.
Through benchmarks and ablations, we confirmed that target networks do indeed slow down training and showed a way to remove them without sacrificing training stability.
We also showed that BatchNorm has the same accelerating effect on training in Deep RL as it does in supervised deep learning.
The combined effect of removing target networks and adding BatchNorm is what makes \CrossQ{} so efficient.
We investigated the relationship between the Q estimation bias and the learning performance of \CrossQ{}, but did not identify a straightforward dependence. This indicates that the relationship between the Q estimation bias and the agent performance is more complex than previously thought.

In future work, it would be interesting to analyze the Q estimation bias more extensively, similar to~\citet{li2022efficient}.
Furthermore, a deeper theoretical analysis of the used BatchNorm approach in the context of RL would be valuable, akin to the works in supervised learning, e.g.,~\citet{Summers2020Four}.
Although the wider critic networks do provide an additional performance boost, they increase the computation cost, which could potentially be reduced.
Finally, while our work focuses on the standard continuous control benchmarking environments, a logical extension would be applying \CrossQ{} to a real robot system and using visual observations in addition to the robot state.
Techniques from image-based RL, such as state augmentation~\citep{laskin2020rad,yarats2021drqv2} and auxiliary losses~\citep{schwarzer2020spr,he2022a2ls}, also aim to learn efficiently from limited data. We believe some of these ideas could potentially be applied to CrossQ.

\subsubsection*{Acknowledgments}
We acknowledge the grant ``Einrichtung eines Labors des Deutschen Forschungszentrum f\"ur K\"unstliche Intelligenz (DFKI) 
an der Technischen Universit\"at Darmstadt" of the Hessisches Ministerium f\"ur Wissenschaft und Kunst. 
This research was also supported by the Research Clusters “The Adaptive Mind” and “Third Wave of AI”, funded by the Excellence Program of the Hessian Ministry of Higher Education, Science, Research and the Arts, Hessian.AI and by the German Research Foundation (DFG): 417962828.

\bibliography{iclr2024_conference}

\begin{thebibliography}{41}
\providecommand{\natexlab}[1]{#1}
\providecommand{\url}[1]{\texttt{#1}}
\expandafter\ifx\csname urlstyle\endcsname\relax
  \providecommand{\doi}[1]{doi: #1}\else
  \providecommand{\doi}{doi: \begingroup \urlstyle{rm}\Url}\fi

\bibitem[Agarwal et~al.(2021)Agarwal, Schwarzer, Castro, Courville, and Bellemare]{agarwal2021rliable}
Rishabh Agarwal, Max Schwarzer, Pablo~Samuel Castro, Aaron Courville, and Marc~G Bellemare.
\newblock Deep reinforcement learning at the edge of the statistical precipice.
\newblock In \emph{Advances in neural information processing systems}, 2021.

\bibitem[Ba et~al.(2016)Ba, Kiros, and Hinton]{ba2016layernorm}
Jimmy~Lei Ba, Jamie~Ryan Kiros, and Geoffrey~E Hinton.
\newblock Layer normalization.
\newblock \emph{arXiv preprint arXiv:1607.06450}, 2016.

\bibitem[Chen et~al.(2021)Chen, Wang, Zhou, and Ross]{chen2021redq}
Xinyue Chen, Che Wang, Zijian Zhou, and Keith Ross.
\newblock Randomized ensembled double {Q}-learning: Learning fast without a model.
\newblock In \emph{International conference on learning representations}, 2021.

\bibitem[Fan et~al.(2020)Fan, Wang, Xie, and Yang]{fan2020theoretical}
Jianqing Fan, Zhaoran Wang, Yuchen Xie, and Zhuoran Yang.
\newblock A theoretical analysis of deep {Q}-learning.
\newblock In \emph{Learning for dynamics and control}, 2020.

\bibitem[Fujimoto et~al.(2018)Fujimoto, Hoof, and Meger]{fujimoto2018td3}
Scott Fujimoto, Herke Hoof, and David Meger.
\newblock Addressing function approximation error in actor-critic methods.
\newblock In \emph{International conference on machine learning}, 2018.

\bibitem[Haarnoja et~al.(2018{\natexlab{a}})Haarnoja, Zhou, Abbeel, and Levine]{haarnoja2018sac}
Tuomas Haarnoja, Aurick Zhou, Pieter Abbeel, and Sergey Levine.
\newblock Soft actor-critic: Off-policy maximum entropy deep reinforcement learning with a stochastic actor.
\newblock In \emph{International conference on machine learning}, 2018{\natexlab{a}}.

\bibitem[Haarnoja et~al.(2018{\natexlab{b}})Haarnoja, Zhou, Hartikainen, Tucker, Ha, Tan, Kumar, Zhu, Gupta, Abbeel, et~al.]{haarnoja2018soft}
Tuomas Haarnoja, Aurick Zhou, Kristian Hartikainen, George Tucker, Sehoon Ha, Jie Tan, Vikash Kumar, Henry Zhu, Abhishek Gupta, Pieter Abbeel, et~al.
\newblock Soft actor-critic algorithms and applications.
\newblock \emph{arXiv preprint arXiv:1812.05905}, 2018{\natexlab{b}}.

\bibitem[Hasselt(2010)]{hasselt2010double}
Hado Hasselt.
\newblock Double {Q}-learning.
\newblock In \emph{Advances in neural information processing systems}, 2010.

\bibitem[He et~al.(2016)He, Zhang, Ren, and Sun]{he2016resnet}
Kaiming He, Xiangyu Zhang, Shaoqing Ren, and Jian Sun.
\newblock Deep residual learning for image recognition.
\newblock In \emph{Conference on computer vision and pattern recognition}, 2016.

\bibitem[He et~al.(2022)He, Zhang, Ren, Liu, Wang, Zhang, Yang, and Li]{he2022a2ls}
Tairan He, Yuge Zhang, Kan Ren, Minghuan Liu, Che Wang, Weinan Zhang, Yuqing Yang, and Dongsheng Li.
\newblock Reinforcement learning with automated auxiliary loss search.
\newblock In \emph{Advances in neural information processing systems}, 2022.

\bibitem[Hiraoka et~al.(2021)Hiraoka, Imagawa, Hashimoto, Onishi, and Tsuruoka]{hiraoka2021droq}
Takuya Hiraoka, Takahisa Imagawa, Taisei Hashimoto, Takashi Onishi, and Yoshimasa Tsuruoka.
\newblock Dropout q-functions for doubly efficient reinforcement learning.
\newblock In \emph{International conference on learning representations}, 2021.

\bibitem[Ioffe(2017)]{ioffe2017batchRenorm}
Sergey Ioffe.
\newblock Batch renormalization: Towards reducing minibatch dependence in batch-normalized models.
\newblock In \emph{Advances in neural information processing systems}, 2017.

\bibitem[Ioffe \& Szegedy(2015)Ioffe and Szegedy]{ioffe2015batchnorm}
Sergey Ioffe and Christian Szegedy.
\newblock Batch normalization: Accelerating deep network training by reducing internal covariate shift.
\newblock In \emph{International conference on machine learning}, 2015.

\bibitem[Janner et~al.(2019)Janner, Fu, Zhang, and Levine]{janner2019mbpo}
Michael Janner, Justin Fu, Marvin Zhang, and Sergey Levine.
\newblock When to trust your model: Model-based policy optimization.
\newblock In \emph{Advances in neural information processing systems}, 2019.

\bibitem[Kim et~al.(2019)Kim, Asadi, Littman, and Konidaris]{Kim2019DeepMellowRT}
Seungchan Kim, Kavosh Asadi, Michael~L. Littman, and George~Dimitri Konidaris.
\newblock Deepmellow: Removing the need for a target network in deep {Q}-learning.
\newblock In \emph{International joint conference on artificial intelligence}, 2019.

\bibitem[Kingma \& Ba(2015)Kingma and Ba]{Kingma2014AdamAM}
Diederik~P. Kingma and Jimmy Ba.
\newblock Adam: {A} method for stochastic optimization.
\newblock In \emph{International conference on learning representations}, 2015.

\bibitem[Laskin et~al.(2020)Laskin, Lee, Stooke, Pinto, Abbeel, and Srinivas]{laskin2020rad}
Misha Laskin, Kimin Lee, Adam Stooke, Lerrel Pinto, Pieter Abbeel, and Aravind Srinivas.
\newblock Reinforcement learning with augmented data.
\newblock In \emph{Advances in neural information processing systems}, 2020.

\bibitem[Li et~al.(2022)Li, Kumar, Kostrikov, and Levine]{li2022efficient}
Qiyang Li, Aviral Kumar, Ilya Kostrikov, and Sergey Levine.
\newblock Efficient deep reinforcement learning requires regulating overfitting.
\newblock In \emph{International conference on learning representations}, 2022.

\bibitem[Lillicrap et~al.(2016)Lillicrap, Hunt, Pritzel, Heess, Erez, Tassa, Silver, and Wierstra]{lillicrap2016ddpg}
Timothy~P Lillicrap, Jonathan~J Hunt, Alexander Pritzel, Nicolas Heess, Tom Erez, Yuval Tassa, David Silver, and Daan Wierstra.
\newblock Continuous control with deep reinforcement learning.
\newblock In \emph{International conference on machine learning}, 2016.

\bibitem[Lim et~al.(2023)Lim, Kim, Choo, and Choi]{lim2023ttn}
Hyesu Lim, Byeonggeun Kim, Jaegul Choo, and Sungha Choi.
\newblock {TTN}: A domain-shift aware batch normalization in test-time adaptation.
\newblock In \emph{International conference on learning representations}, 2023.

\bibitem[Mnih et~al.(2015)Mnih, Kavukcuoglu, Silver, Rusu, Veness, Bellemare, Graves, Riedmiller, Fidjeland, Ostrovski, et~al.]{mnih2015dqn}
Volodymyr Mnih, Koray Kavukcuoglu, David Silver, Andrei~A Rusu, Joel Veness, Marc~G Bellemare, Alex Graves, Martin Riedmiller, Andreas~K Fidjeland, Georg Ostrovski, et~al.
\newblock Human-level control through deep reinforcement learning.
\newblock \emph{Nature}, 518\penalty0 (7540):\penalty0 529--533, 2015.

\bibitem[Morales(2020)]{morales2020grokking}
M.~Morales.
\newblock \emph{Grokking deep reinforcement learning}.
\newblock Manning Publications, 2020.

\bibitem[Nikishin et~al.(2022)Nikishin, Schwarzer, D’Oro, Bacon, and Courville]{nikishin2022primacy}
Evgenii Nikishin, Max Schwarzer, Pierluca D’Oro, Pierre-Luc Bacon, and Aaron Courville.
\newblock The primacy bias in deep reinforcement learning.
\newblock In \emph{International conference on machine learning}, 2022.

\bibitem[Ota et~al.(2020)Ota, Oiki, Jha, Mariyama, and Nikovski]{ota2020can}
Kei Ota, Tomoaki Oiki, Devesh Jha, Toshisada Mariyama, and Daniel Nikovski.
\newblock Can increasing input dimensionality improve deep reinforcement learning?
\newblock In \emph{International conference on machine learning}, 2020.

\bibitem[Ota et~al.(2021)Ota, Jha, and Kanezaki]{ota2021widenets}
Kei Ota, Devesh~K Jha, and Asako Kanezaki.
\newblock Training larger networks for deep reinforcement learning.
\newblock \emph{arXiv preprint arXiv:2102.07920}, 2021.

\bibitem[Pham et~al.(2022)Pham, Liu, and HOI]{pham2022continual}
Quang Pham, Chenghao Liu, and Steven HOI.
\newblock Continual normalization: Rethinking batch normalization for online continual learning.
\newblock In \emph{International conference on learning representations}, 2022.

\bibitem[Plappert et~al.(2018)Plappert, Andrychowicz, Ray, McGrew, Baker, Powell, Schneider, Tobin, Chociej, Welinder, et~al.]{plappert2018multi}
Matthias Plappert, Marcin Andrychowicz, Alex Ray, Bob McGrew, Bowen Baker, Glenn Powell, Jonas Schneider, Josh Tobin, Maciek Chociej, Peter Welinder, et~al.
\newblock Multi-goal reinforcement learning: Challenging robotics environments and request for research.
\newblock \emph{arXiv preprint arXiv:1802.09464}, 2018.

\bibitem[Puterman(2014)]{puterman2014mdp}
Martin~L Puterman.
\newblock \emph{Markov decision processes: {D}iscrete stochastic dynamic programming}.
\newblock John Wiley \& Sons, 2014.

\bibitem[Raffin et~al.(2021)Raffin, Hill, Gleave, Kanervisto, Ernestus, and Dormann]{stable-baselines3}
Antonin Raffin, Ashley Hill, Adam Gleave, Anssi Kanervisto, Maximilian Ernestus, and Noah Dormann.
\newblock Stable-baselines3: Reliable reinforcement learning implementations.
\newblock \emph{Journal of machine learning research}, 22\penalty0 (268):\penalty0 1--8, 2021.

\bibitem[Santurkar et~al.(2018)Santurkar, Tsipras, Ilyas, and Madry]{santurkar2018howbatchnorm}
Shibani Santurkar, Dimitris Tsipras, Andrew Ilyas, and Aleksander Madry.
\newblock How does batch normalization help optimization?
\newblock In \emph{Advances in neural information processing systems}, 2018.

\bibitem[Schwarzer et~al.(2021)Schwarzer, Anand, Goel, Hjelm, Courville, and Bachman]{schwarzer2020spr}
Max Schwarzer, Ankesh Anand, Rishab Goel, R~Devon Hjelm, Aaron Courville, and Philip Bachman.
\newblock Data-efficient reinforcement learning with self-predictive representations.
\newblock In \emph{International conference on learning representations}, 2021.

\bibitem[Singh \& Shrivastava(2019)Singh and Shrivastava]{singh2019evalnorm}
Saurabh Singh and Abhinav Shrivastava.
\newblock Evalnorm: Estimating batch normalization statistics for evaluation.
\newblock In \emph{International conference on computer vision}, 2019.

\bibitem[Smith et~al.(2022)Smith, Kostrikov, and Levine]{smith2022walk}
Laura Smith, Ilya Kostrikov, and Sergey Levine.
\newblock A walk in the park: Learning to walk in 20 minutes with model-free reinforcement learning.
\newblock \emph{arXiv preprint arXiv:2208.07860}, 2022.

\bibitem[Srivastava et~al.(2014)Srivastava, Hinton, Krizhevsky, Sutskever, and Salakhutdinov]{srivastava2014dropout}
Nitish Srivastava, Geoffrey Hinton, Alex Krizhevsky, Ilya Sutskever, and Ruslan Salakhutdinov.
\newblock Dropout: {A} simple way to prevent neural networks from overfitting.
\newblock \emph{Journal of machine learning research}, 15\penalty0 (1):\penalty0 1929--1958, 2014.

\bibitem[Summers \& Dinneen(2020)Summers and Dinneen]{Summers2020Four}
Cecilia Summers and Michael~J. Dinneen.
\newblock Four things everyone should know to improve batch normalization.
\newblock In \emph{International conference on learning representations}, 2020.

\bibitem[Tassa et~al.(2018)Tassa, Doron, Muldal, Erez, Li, Casas, Budden, Abdolmaleki, Merel, Lefrancq, et~al.]{tassa2018dm_control}
Yuval Tassa, Yotam Doron, Alistair Muldal, Tom Erez, Yazhe Li, Diego de~Las Casas, David Budden, Abbas Abdolmaleki, Josh Merel, Andrew Lefrancq, et~al.
\newblock Deepmind control suite.
\newblock \emph{arXiv preprint arXiv:1801.00690}, 2018.

\bibitem[Todorov et~al.(2012)Todorov, Erez, and Tassa]{todorov2012mujoco}
Emanuel Todorov, Tom Erez, and Yuval Tassa.
\newblock Mujoco: A physics engine for model-based control.
\newblock In \emph{International conference on intelligent robots and systems}, 2012.

\bibitem[Towers et~al.(2023)Towers, Terry, Kwiatkowski, Balis, Cola, Deleu, Goulão, Kallinteris, KG, Krimmel, Perez-Vicente, Pierré, Schulhoff, Tai, Shen, and Younis]{towers2023gymnasium}
Mark Towers, Jordan~K. Terry, Ariel Kwiatkowski, John~U. Balis, Gianluca~de Cola, Tristan Deleu, Manuel Goulão, Andreas Kallinteris, Arjun KG, Markus Krimmel, Rodrigo Perez-Vicente, Andrea Pierré, Sander Schulhoff, Jun~Jet Tai, Andrew Tan~Jin Shen, and Omar~G. Younis.
\newblock Gymnasium, 2023.

\bibitem[Yang et~al.(2021)Yang, Ajay, and Agrawal]{yang2021overcoming}
Ge~Yang, Anurag Ajay, and Pulkit Agrawal.
\newblock Overcoming the spectral bias of neural value approximation.
\newblock In \emph{International conference on learning representations}, 2021.

\bibitem[Yarats et~al.(2021)Yarats, Fergus, Lazaric, and Pinto]{yarats2021drqv2}
Denis Yarats, Rob Fergus, Alessandro Lazaric, and Lerrel Pinto.
\newblock Mastering visual continuous control: Improved data-augmented reinforcement learning.
\newblock In \emph{International conference on learning representations}, 2021.

\bibitem[Ziebart et~al.(2008)Ziebart, Maas, Bagnell, Dey, et~al.]{ziebart2008maxent}
Brian~D Ziebart, Andrew~L Maas, J~Andrew Bagnell, Anind~K Dey, et~al.
\newblock Maximum entropy inverse reinforcement learning.
\newblock In \emph{AAAI conference on artificial intelligence}, 2008.

\end{thebibliography}
\bibliographystyle{iclr2024_conference}

\newpage
\appendix
\section{Appendix}
\subsection{DeepMind Control Suite Experiments}
\begin{figure}[h]
    \centering
    \includegraphics[width=\textwidth]{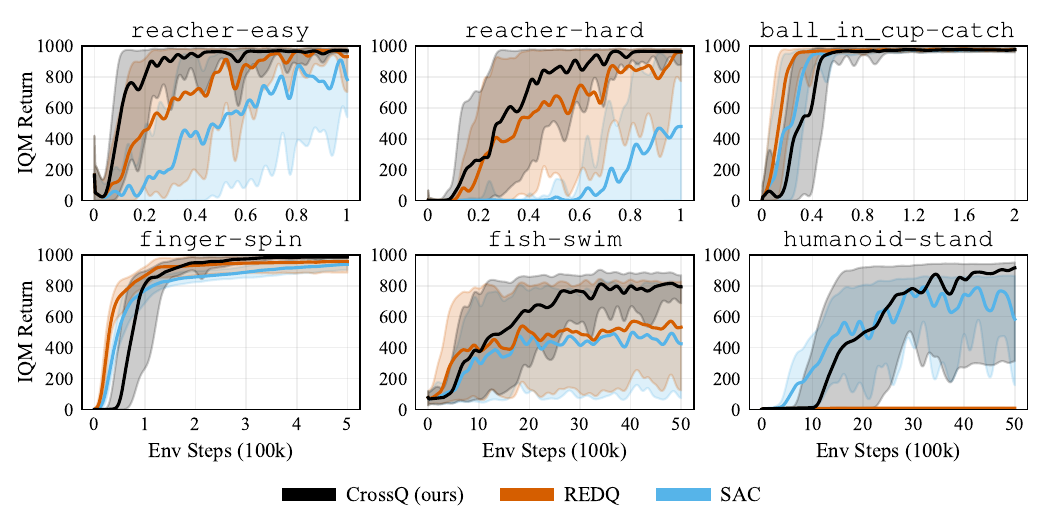}
    \caption{\textbf{Sample efficiency of \CrossQ{} on DeepMind Control.} The experiments here were each performed on $5$ different random seeds. \CrossQ{}'s good sample efficiency transfers well to the presented tasks from the DeepMind Control Suite.}
    \label{fig:sample_efficiency_dmc}
\end{figure}

Figure~\ref{fig:sample_efficiency_dmc} presents an additional set of experiments performed on the DeepMind Control Suite~\citep{tassa2018dm_control}.
The experiments shown here are an extension to the experiments shown in Figure~\ref{ref:sample_efficiency} in the main paper and have been moved to the Appendix due to space constraints.
For the presented tasks, we lowered the learning rate to $8\times10^{-4}$ for all algorithms, and set the \CrossQ{} policy delay to 1. All other hyperparameters remained the same as for the main paper.
\newpage
\subsection{Hyperparameters}

Experiment hyperparameters, used in the main paper. We adapted most hyperparameters that are commonly used in other works~\citep{haarnoja2018soft,chen2021redq,hiraoka2021droq}. \\ The Moving-Average \textit{Momentum} corresponds to $1$ minus the Moving-Average \textit{Update Rate} as defined in both BatchNorm papers~\citep{ioffe2015batchnorm, ioffe2017batchRenorm}. 

\begin{table}[h]
\centering
\caption{Learning Hyperparameters}
\label{tab:hyperparameters}
\begin{tabular}{l|c|c|c|c}
\toprule
\textbf{Parameter}                & SAC  & REDQ & DroQ & \CrossQ{} (ours)\\
\midrule\midrule
Discount Factor ($\gamma$)        & \multicolumn{4}{c}{$0.99$}             \\ \midrule
Learning Rate (Actor \& Critic)      & \multicolumn{4}{c}{$0.001$}            \\ \midrule
Replay Buffer Size                & \multicolumn{4}{c}{$10^6$}           \\\midrule
Batch Size                        & \multicolumn{4}{c}{$256$}               \\\midrule
Activation Function               & \multicolumn{4}{c}{\texttt{relu}}     \\\midrule
Layer Normalization               & \multicolumn{2}{c|}{No} & Yes  & No   \\\midrule
Dropout Rate                      & \multicolumn{2}{c|}{\texttt{N/A}} & $0.01$ & \texttt{N/A}  \\\midrule
BatchNorm / Version              & \multicolumn{3}{c|}{\texttt{N/A}} & \texttt{BRN}
\\\midrule
BatchNorm / Moving-Average Momentum              & \multicolumn{3}{c|}{\texttt{N/A}} &$0.99$
\\\midrule
BatchNorm / \texttt{BRN} Warm-up Steps              & \multicolumn{3}{c|}{\texttt{N/A}} & $10^5$
\\\midrule
Critic Width                      & \multicolumn{3}{c|}{$256$} & $2048$      \\\midrule
Target Update Rate ($\tau$)       & \multicolumn{3}{c|}{$0.005$}         & \texttt{N/A}  \\\midrule
Adam $\beta_1$                    & \multicolumn{3}{c|}{$0.9$}           &  $0.5$  \\\midrule
Update-To-Data ratio (UTD)        & $1$  & \multicolumn{2}{c|}{$20$} & $1$     \\ \midrule
Policy Delay                      & $1$  & \multicolumn{2}{c|}{$20$} & $3$     \\\midrule
Number of  Critics                & $2$  & \multicolumn{1}{c}{$10$} & \multicolumn{2}{|c}{$2$} \\
\bottomrule
\end{tabular}
\end{table}

\subsection{Wallclock Time Measurement}
\label{sec:wallclock}
Wallclock times were measured by timing and averaging over four seeds each and represent \textit{pure training times}, without the overhead of synchronous evaluation and logging, until reaching $5\times10^6$ environment steps. The times are recorded on an \texttt{Nvidia RTX 3090 Turbo} with an \texttt{AMD EPYC 7453} CPU. 

\begin{table}[b]
    \centering
    \vspace{-1.5em}
    \caption{\textbf{Wallclock times.} Evaluated for \CrossQ{} and baselines across environments in hours and recorded on an \texttt{RTX 3090}, the details of the measurement procedure are described in Appendix~\ref{sec:computational_efficiency}.
    Comparing \CrossQ{} with \CrossQ{} (Small) and SAC, it is apparent that using wider critic networks does come with a performance penalty.
    However, compared to REDQ and DroQ, one clearly sees the substantial improvement in Wallclock time of \CrossQ{} over those baselines.
    }
    \vspace{.8em}
    \begin{tabular}{ lccccc  }
    \toprule
    & \multicolumn{5}{c}{Wallclock Time [hours]} \\
    & SAC & \CrossQ{} (small) & \textbf{\CrossQ{} (ours)}  & REDQ & DroQ \\
    \midrule
    \texttt{HumanoidStandup-v4}   & 1.5 & 2.1 & 2.2 & 8.7 &  7.5  \\
    \texttt{Walker2d-v4}          & 0.9 & 0.9 & 1.1 & 4.0 &  4.1  \\
    \texttt{Ant-v4}               & 0.9 & 1.2 & 1.5 & 4.7 &  4.7  \\
    \texttt{HalfCheetah-v4}       & 0.8 & 1.2 & 1.5 & 4.1 &  4.4  \\
    \texttt{Hopper-v4}            & 1.0 & 1.1 & 1.3 & 4.1 &  4.2  \\
    \bottomrule
    \end{tabular}
    \label{tab:computation_times}
\end{table}

\newpage

\subsection{Evolving Action Distributions}
\begin{figure}[h]
    \centering
    \includegraphics[width=\textwidth]{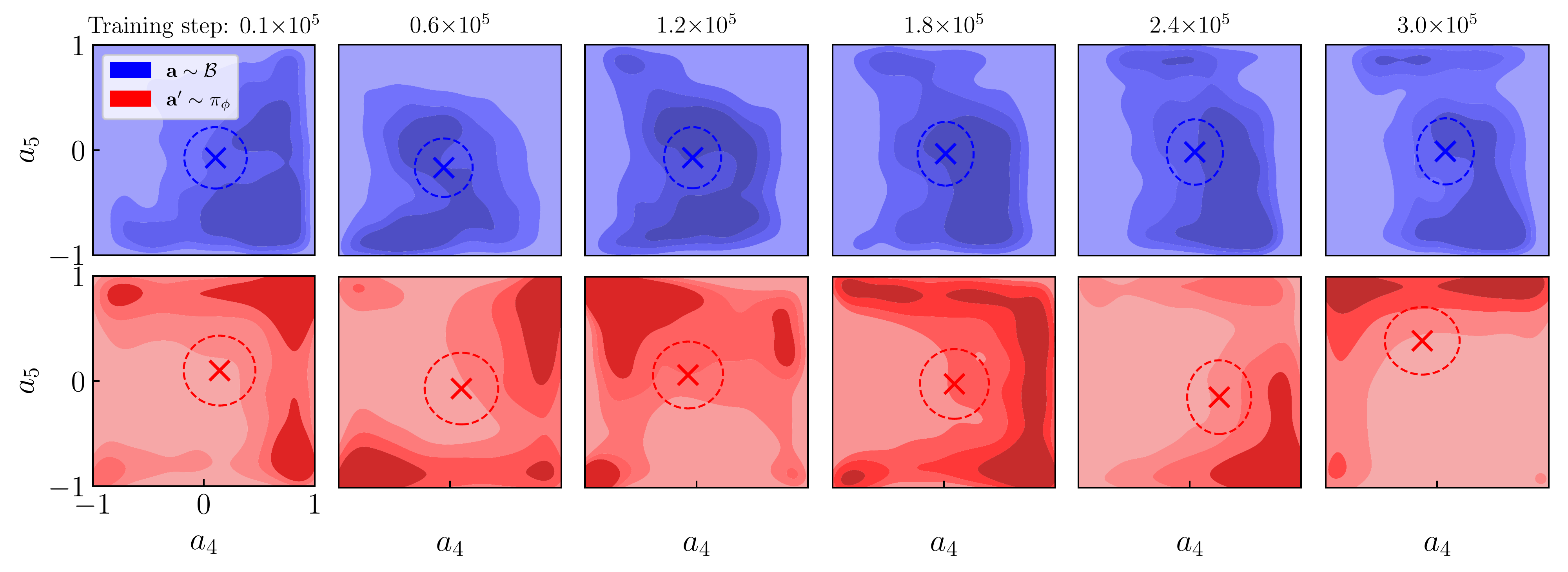}
    \caption{\textbf{Replay and policy action distributions are different, and evolve during training.} We train an agent for $300,000$ steps on \texttt{Walker2d}. We take snapshots of the replay buffer $\mathcal{B}$ and policy $\pi_\mathbf{\phi}$ every $60,000$ steps. For each snapshot (one column), we sample a large batch of $10,000$ transitions $(\vs, \va, \vs', \va'=\pi_\mathbf{\phi}(\vs'))$ and use this to compute a visually interpretable 2D kernel density estimate of the distributions of $\va$ (\textcolor{blue}{blue}) and $\va'$ (\textcolor{red}{red}), as seen through the action-space dimensions $4$ and $5$. The cross denotes the mean, and the dashed ellipse is one standard deviation wide for each of the two dimensions.
    We observe that the distributions as well as the means and standard deviations of the off-policy and on-policy actions are visibly and persistently different throughout the training run, and keep drifting as the training progresses. This discrepancy implies that BatchNorm must be used with care in off-policy TD learning.}
    \label{fig:evolving_action_densities}
\end{figure}

\newpage
\subsubsection{Ablating the Different Design Choices and Hyperparameters}

\begin{figure}[h]
    \centering
    \includegraphics[width=\textwidth]{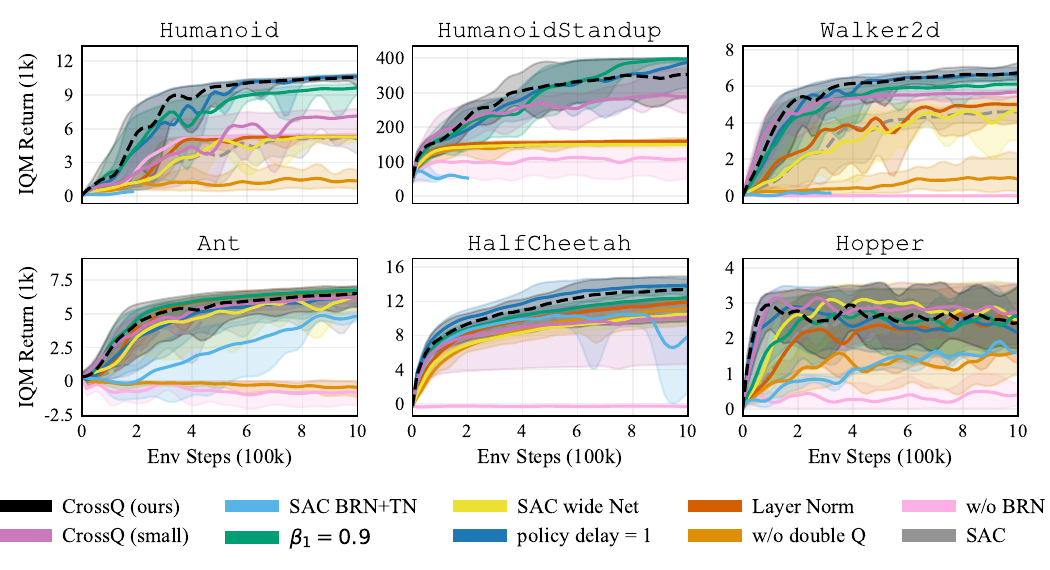}
    \vspace{-2em}
    \caption{\textbf{\CrossQ{} ablation study.} We ablate across different hyperparameter settings and architectural configurations.
    Using the same network width as SAC, \textcolor{fmagenta}{\CrossQ{} (small)} shows weaker performance, yet is still competitive with \CrossQ{} in four out of six environments. At the same time, \textcolor{fyellow}{SAC with a wider critic} does not work better. Using the default Adam momentum \textcolor{fgreen}{$\beta_1=0.9$} instead of $0.5$ degrades performance in some environments. Using a \textcolor{fblue}{policy delay of $1$} instead of $3$ has a very small effect, except on $\texttt{Ant}$. Using \textcolor{foccer}{LayerNorm} instead of BatchNorm results in slower learning; it also trains stably without target networks. \textcolor{fpink}{Removing BatchNorm} results in failure of training due to divergence. \textcolor{cyan}{Adding BatchNorm to SAC} and reusing the live critic's normalization moments in the target network fails to train. Training \textcolor{fred}{without double Q} networks (single critic) harms performance.
    \label{fig:crossq_ablations}}
    \vspace{-1em}
\end{figure}

Figure \ref{fig:crossq_ablations} depicts in detail the \CrossQ{} and SAC ablations, previously shown in aggregate form by Figure \ref{fig:bar_ablations}.

\newpage
\subsection{REDQ and DroQ Ablations}
\begin{figure}[h]
    \centering
    \vspace{-1em}
    \includegraphics[width=\textwidth]{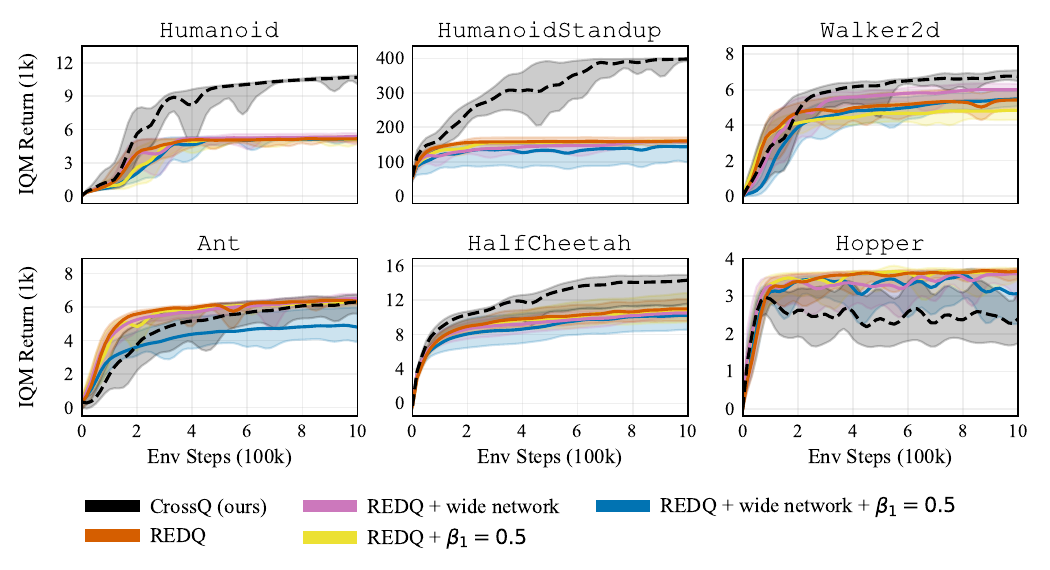}
    \vspace{-2em}
    \caption{\textbf{REDQ ablation.} Showing performance for different combinations of the \CrossQ{} hyperparameters. The changes in hyperparameters do not help REDQ to get better performance. In fact, in some cases, they even hurt the performance.}
    \label{fig:redq_hyperparameter_ablation}
\end{figure}

\begin{figure}[h]
    \centering
    \vspace{-1em}
    \includegraphics[width=\textwidth]{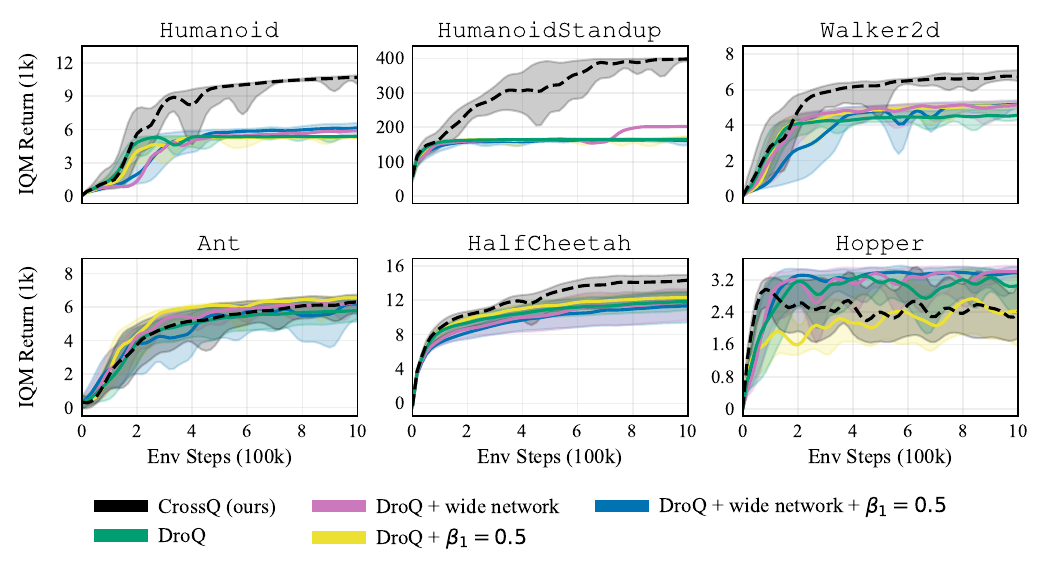}
    \vspace{-2em}
    \caption{\textbf{DroQ ablation.} The changes in hyperparameters do not help DroQ to get better performance overall. In \texttt{Hopper} and \texttt{Ant}, performance rises to the \CrossQ{} performance, however, on the \texttt{Humanoid}, it hurts performance.}
    \label{fig:droq_hyperparameter_ablation}
\end{figure}

Figures~\ref{fig:redq_hyperparameter_ablation} and~\ref{fig:droq_hyperparameter_ablation} show REDQ and DroQ ablations on $5$ seeds each. They show both baselines with the \CrossQ{} hyperparameters: wider critic networks as well as $\beta_1=0.5$.
Neither baseline benefits from the added changes. In most cases, the performance is unchanged, while in some cases, it deteriorates.
The dashed black line shows \CrossQ{} as a reference.

\newpage
\subsection{Effect of Activations and Normalizers on Learning Stability}
\label{sec:diverse_activations}
Figure~\ref{fig:bounded_activations} depicts a small exploratory experiment in which we remove target networks from SAC, and train it with different activation functions and feature normalizers. We do this only to explore whether the boundedness of activations has an influence on training stability. We learn from this experiment that SAC with $\mathrm{tanh}$ activations trains without divergence, allowing us to conduct the study in Section~\ref{sec:bounded_activations}. We also observe that at least two feature normalization schemes (on top of the unbounded relu activations) permit divergence-free optimization.

For vectors $\vx$, $\mathrm{relu\_over\_max}(\vx)$ denotes a simple normalization scheme using an underlying unbounded activation: $\mathrm{relu}(\vx)/\mathrm{max}(\vx)$, with the maximum computed over the entire feature vector. $\mathrm{layernormed\_relu}$ simply denotes LayerNorm applied \textit{after} the $\mathrm{relu}$ activations. Both of these schemes prevent divergence. Using LayerNorm \textit{before} the $\mathrm{relu}$ activations also prevent divergence, and is already explored in the ablations in Figure~\ref{fig:crossq_ablations}. None of these normalizers perform as strongly as BatchNorm. 

A thorough theoretical or experimental study of how activations and normalizers affect the stability of Deep RL is beyond the scope of this paper. We hope, however, that our observations help inform future research directions for those interested in this topic.

\begin{figure}[h]
    \centering
    \includegraphics[width=\textwidth]{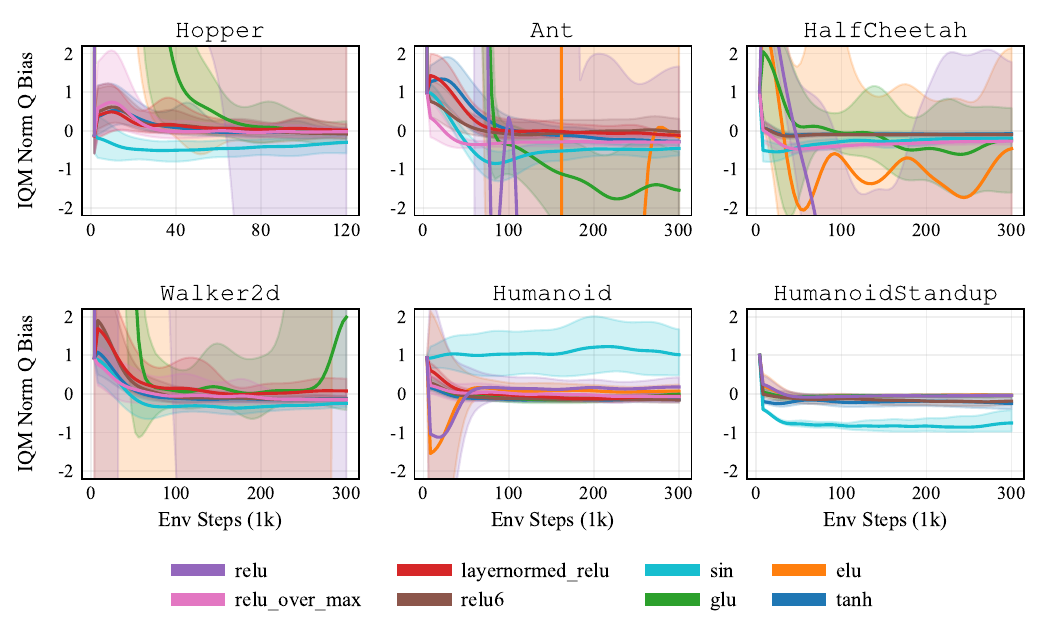}
    \caption{\textbf{(In)stability of SAC without target networks.} Observed through the Q estimation bias. In this small-scale experiment, we run SAC with unbounded ($\mathrm{relu, glu, elu}$) and bounded ($\mathrm{tanh, relu6, sin}$) activation functions, as well as ``indirectly" bounded activations through the use of two custom normalizers other than BatchNorm ($\mathrm{relu\_over\_max, layernormed\_relu}$). SAC variants with unbounded activations appear highly unstable in most environments, whereas the variants with bounded activations (as well as the normalizers) do not diverge, maintaining relatively low bias.
    }
    \label{fig:diverse_activations}
\end{figure}

\newpage
\subsection{Normalized $Q$ Bias Plots}
\label{app:q_bias}

Figure~\ref{fig:q_bias} shows the results of the Q function bias analysis for all environments.

\begin{figure}[h]
    \centering
    \includegraphics[width=\textwidth]{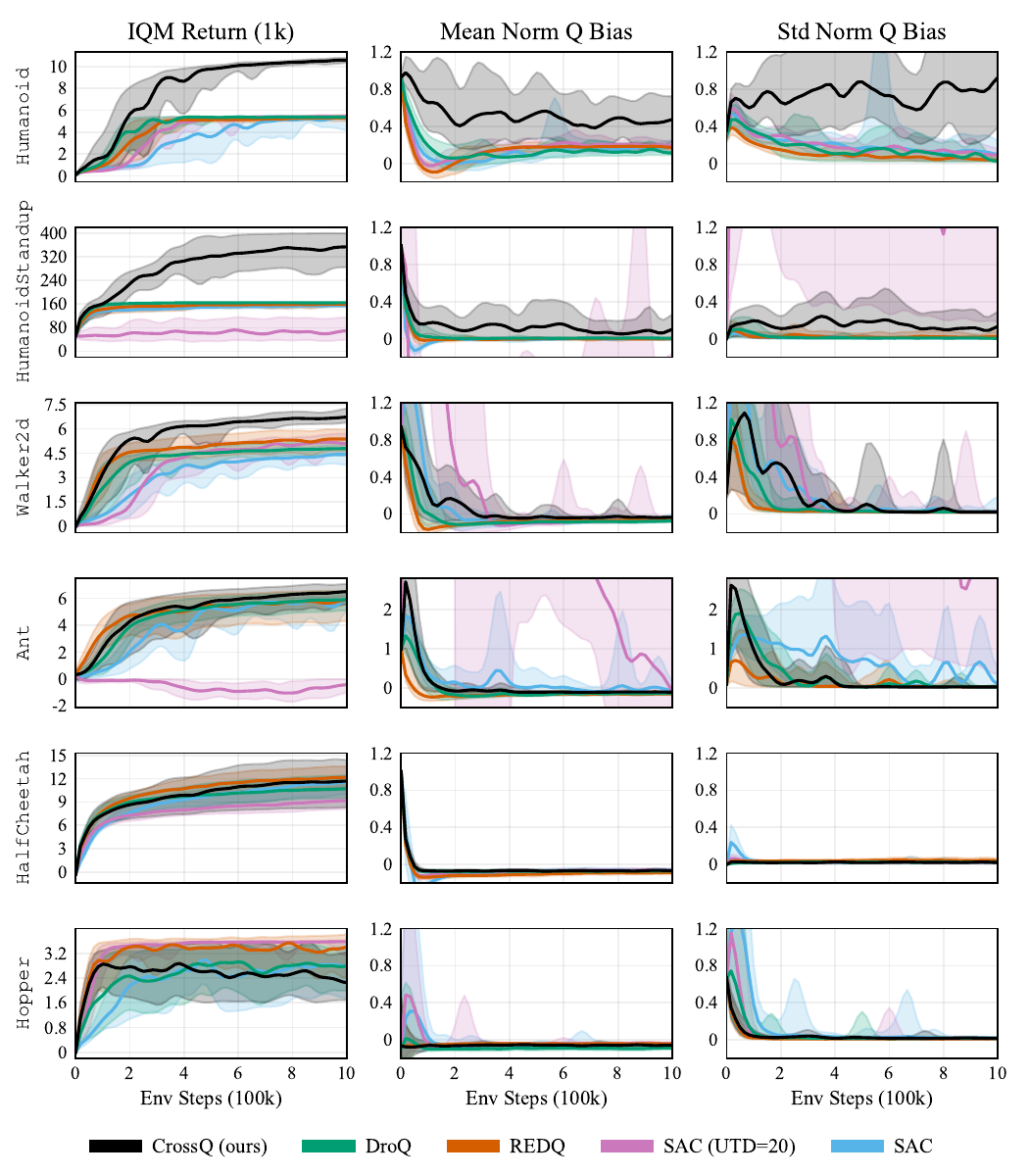}
    \caption{\textbf{Q estimation bias.} Mean and standard deviation of the normalized Q function bias, computed as described by~\citet{chen2021redq}.
    As in the main paper, we do not find a straightforward connection between normalized Q function bias and learning performance.
    \CrossQ{} generally shows the same or larger Q estimation bias compared to REDQ but matches or outperforms REDQ in learning speed, especially on the challenging \texttt{Humanoid} tasks.
    }
    \label{fig:q_bias}
\end{figure}

\end{document}